\pdfoutput=1

\documentclass[11pt]{article}

\usepackage[]{acl}
\usepackage{times}
\usepackage{latexsym}
\usepackage[T1]{fontenc}
\usepackage[utf8]{inputenc}
\usepackage{microtype}
\usepackage{graphicx}
\usepackage{amssymb}
\usepackage{hyperref}
\usepackage{multirow}
\usepackage{mathtools}
\usepackage{float}
\usepackage{xurl}

\title{Persian Emotion Detection using ParsBERT and Imbalanced Data Handling Approaches}

\author{
  Amirhossein Abaskohi\textsuperscript{1},
  Nazanin Sabri\textsuperscript{2},
  Behnam Bahrak\textsuperscript{1}\\
  \textsuperscript{1}School of Electrical and Computer Engineering \\ College of Engineering, University of Tehran, Tehran, Iran \\
  {\texttt{\{amir.abaskohi, bahrak\}@ut.ac.ir}} \\
  \textsuperscript{2}School of Computer Science
  University of California, San Diego, La Jolla, CA 92093 \\
  {\texttt{nsabri@ucsd.edu}}
}

\begin{document}
\maketitle
\begin{abstract}
\label{sec:abstract}

Emotion recognition is one of the machine learning applications which can be done using text, speech, or image data gathered from social media spaces. Detecting emotion can help us in different fields, including opinion mining. With the spread of social media, different platforms like Twitter have become data sources, and the language used in these platforms is informal, making the emotion detection task difficult. EmoPars and ArmanEmo are two new human-labeled emotion datasets for the Persian language. These datasets, especially EmoPars, are suffering from inequality between several samples between two classes. In this paper, we evaluate EmoPars and compare them with ArmanEmo. Throughout this analysis, we use data augmentation techniques, data re-sampling, and class-weights with Transformer-based Pretrained Language Models(PLMs) to handle the imbalance problem of these datasets. Moreover, feature selection is used to enhance the models' performance by emphasizing the text's specific features. In addition, we provide a new policy for selecting data from EmoPars, which selects the high-confidence samples; as a result, the model does not see samples that do not have specific emotion during training. Our model reaches a Macro-averaged F1-score of 0.81 and 0.76 on ArmanEmo and EmoPars, respectively, which are new state-of-the-art results in these benchmarks.\footnote{Code is publicly available at \url{https://github.com/AmirAbaskohi/Persian-Emotion-Detection-using-ParsBERT-and-Imbalanced-Data-Handling-Approaches}}

\end{abstract}

\section{Introduction}
\label{sec:introduction}

In people's social and professional lives, emotional expression and detection are crucial due to the fact that systems can change their responses and behavioral patterns in response to human emotions, improving the naturalness of interactions\cite{tzirakis2019real}. They strongly influence the variety and quality of our experiences and social interactions, and they are intimately tied to our cognitive and communicative abilities\cite{dolan2002emotion}. Social networks, for example, have evolved into mediums for users to express their emotions\cite{tettegah2015emotions}. Additionally, they are valuable tools to keep up with the news. Furthermore, finding new social trends based on stakeholder groups' or the general public's expectations, attitudes, and dispositions is the aim of opinion research. Opinion research is often used in policymaking to more accurately predict the effects of proposed changes and to better communicate the projected advantages and drawbacks\cite{sobkowicz2012opinion}. As a result, we use the data from these platforms for emotion recognition for opinion mining.

Sentiment analysis tools are helpful if we just want to check the polarity of a sentence or document\cite{andalibi2020human}. In this paper, we want to go deeper, trying to detect a social media post's emotion since sentiment analysis can find whether a sample is positive or not, whereas emotion is a more complex concept. With emotion recognition, we can get more granular data, bringing out the real emotions of the user who created that specific post.

The precision with which people can gauge the emotions of others varies greatly\cite{gendron2014perceptions} and this is a reason why automated systems have problems in this task. Automating the identification of facial emotions from video, spoken expressions from audio, written expressions from text, and physiology, as measured by wearables, has received the most attention to date. The technology generally performs best when it integrates several modalities into the environment\cite{tzirakis2017end}. However, gathering enough synchronous multi-modal data is not easy, and as a result, there is a need for improving single-modal models. Here our focus is using text data for emotion recognition.

One of the usages of emotion recognition is in detecting psychological disorders. Severe social interaction deficiencies are a hallmark of borderline personality disorder (BPD), which may be related to abnormalities in emotion perception\cite{minzenberg2006social, domes2009emotion}. Furthermore, emotion recognition is essential for society's satisfaction evaluation. This evaluation can be used in different situations like an election or even a recent pandemic, COVID-19\cite{esmin2012hierarchical, shyry2020election, melendez2020emotion}. Moreover, emotion recognition can be used in examining the well-being of people under different circumstances. For instance, in \cite{schlegel2021emotion}, emotion recognition was used as a predictor of well-being during the COVID-19 pandemic. Consequently, emotion detection is a helpful research line, and using social media, like Twitter, whose data type is textual data, plays an important role here since they provide a massive amount which makes the data collection pipeline a lot easier. However, the mentioned papers are in English, and in this paper, we propose an emotion recognition model to use in the areas above. Despite several pieces of research in English, in Persian, mining emotions from text and other media needs more research as the provided datasets for this task in Persian is no specific model for these datasets is provided. EmoPars\cite{sabri-etal-2021-emopars-collection} and ArmanEmo\cite{mirzaee2022armanemo} are two famous textual emotion datasets. Although some methods for detecting emotions are provided in \cite{mirzaee2022armanemo}, in this paper, we try to improve those approaches by solving the imbalance dataset issue and providing a feature-selecting method.

In order to achieve a high F1-score in Persian emotion detection, we attempt to use transformer architecture\cite{vaswani2017attention} and various methods for handling imbalanced data such as undersampling for the major class, suitable class weights, F1-Cross-Entropy loss function, and data augmentation techniques. Our method reaches a state-of-the-art result in both EmoPars\cite{sabri-etal-2021-emopars-collection} and ArmanEmo\cite{mirzaee2022armanemo} benchmarks. In general, the main contributions of this paper are:

\begin{itemize}
    \item {
        Introducing new data sampling on the EmoPars dataset to improve the results on this dataset. Our strategy aims to use more data in contrast to the methodology presented in \cite{mirzaee2022armanemo}, which eliminates 96\% of the tweets in this dataset.
    }
    \item {
        Improving the model's fairness in EmoPars benchmark by using random undersampling for major class and Easy Data Augmentation(synonym replacement, word addition, word elimination, and shuffling)to decrease the differences in the number of samples in classes and using proper class weights and F1-Cross-Entropy loss function.
    }
    \item {
        Introducing a feature extraction method to help the model to pay more attention to specific features in the data, which has much information about the emotion of the sample, such as hashtags and emojis.
    }
\end{itemize}

\section{Related Works}
\label{sec:related-work}

In this section, we first discuss emotion detection studies that have been previously investigated, especially in Persian texts. We then provide a brief review of approaches for handling imbalanced datasets. Finally, we will look at the feature extraction approaches in English and Persian texts.

\subsection{Emotion Detection Methods In Texts}
\label{subsec:emotionmethods}

People still prefer using text to communicate their thoughts regarding other people, objects, or events\cite{warsi2021text}. Due to the nature of the data, it is still more challenging to extract emotion from the text since the whole text, including metaphors and sarcasm, should be considered\cite{skorzewski2022using}. Sometimes several emotions might be present in a single paragraph. Additionally, some words in texts have many meanings, while others express the same feeling with multiple words.

There are several methods for detecting textual emotions. Affective computing includes tasks like emotion recognition, and diverse academics have divided the computational techniques employed in this field into several categories. These researchers' methodologies may be broadly divided into four groups: machine learning method, lexicon-based method, keyword-based method, and hybrid method\cite{andalibi2020human}. Machine learning methods got more popular due to the adequate performance of word embedding and transformer architecture in sentiment analysis tasks, and this was not just in English. 

Emotion recognition in English has been investigated in several studies. In speech emotion recognition, various approaches, including self-supervised, multilingual fusion, and knowledge transfer, were used\cite{neumann2018cross, sarkar2020self, iosifov2022transferability}. In text emotion recognition, using the psychological frameworks offered by \cite{russell1980circumplex} and \cite{ekman1992argument} to map documents, phrases, and words to a set of emotions was among the first approaches. The problem with traditional methodologies was then overcome by machine learning-based approaches, leading to notable improvements in text emotional recognition\cite{zhang2016grasp}. In text analysis research, deep learning-based methods have emerged as the cutting edge. They have demonstrated the capacity to pick up on the key emotion characteristics, which has significantly improved text emotion identification performance\cite{alswaidan2020survey}. Convolutional Neural Networks(CNNs) and Long Short Term Memories(LSTMs) are among these approaches\cite{Shrivastava2019, mahto2022hierarchical, alla2022emotion}. With the introduction of Transformer architecture, transformer-based, especially BERT-based models, were used for emotion recognition on text data and multi-modal scenarios\cite{li2020hierarchical, acheampong2021transformer}.

Although emotion recognition in Persian has been studied in speech and text domains\cite{hamidi2012emotion, pourebrahim2021parallel, yazdani2021emotion}, very few studies have been conducted on emotion detection from Persian text\cite{sadeghi2021automatic}. One of the reasons for this is the few Persian datasets available for emotion detection. After the introduction of EmoPars\cite{sabri-etal-2021-emopars-collection}, the study of this topic has started with more effort. This dataset includes Persian Tweets and will be introduced in Section \ref{subsec:dataset}. However, emotion detection is not investigated in the provided paper, and just the data and its statistics are provided. ArmanEmo\cite{mirzaee2022armanemo} is another dataset that includes other data sources like Instagram. In addition to introducing a new dataset, they performed transformer-based approaches including ParsBERT\cite{ParsBERT} and XLM-RoBERTa\cite{conneau-etal-2020-unsupervised} to evaluate their dataset and EmoPars dataset. These models will be used in this paper as well. Moreover, we will investigate different approaches for solving the imbalance problem of these datasets and improving the performance of the model as well.

In Persian emotion detection, Sadeghi et al.\cite{sadeghi2021automatic} used a hybrid approach by combining the cognitive features and Word2Vec\cite{mikolov2013efficient} embedding. The emotional construction, keywords, and parts of speech are utilized to construct the cognitive features. Then the features are passed to a Gated Recurrent Unit (GRU)\cite{cho2014properties} network for classification. However, this method needs human supervision which makes it hard to use in other scenarios in this paper our goal is to have a more automatic approach. Since we are dealing with imbalanced data sets illustrated in Section \ref{subsec:dataset}, we need to refer to previous work on how to deal with these datasets.

\subsection{Handling Imbalanced Datasets}
\label{subsec:imbalancedhandling}

Unbalanced datasets have become increasingly important in machine learning in recent years since if the data set is imbalanced, then you get a high accuracy just by predicting the majority class. However, you fail to capture the minority class. Hence, the model does not give good results in production and unseen data since it would completely ignore the minority class in favor of the majority class. The ratio between the majority and minority classes can range from 100:1 to 1000:1\cite{makki2019experimental}, meaning that more people belong to the majority class than the minority class. As a consequence, some helpful information about the data itself, which could be necessary for building rule-based classifiers, may be lost. 

In data-level approaches, Majority Weighted Minority Oversampling Technique (MWMOTE)\cite{barua2012mwmote} has been introduced to use oversampling and weights for major classes to solve the negative effect of imbalanced data in 20 real-world datasets. In another study, Neighbourhood Balanced Bagging (NBBag)\cite{blaszczynski2015neighbourhood} was introduced. In this approach, sampling probabilities of examples are modified according to the class distribution in their neighborhood. Two versions are considered: the first one keeps a larger size of bootstrap samples by hybrid over-sampling, and the other reduces this size with stronger under-sampling.

In algorithm-level approaches, defining class weights for machine learning models has been used in random forest algorithms to overcome imbalanced data problems in medical data\cite{zhu2018class}. Furthermore, different loss functions have been used to increase the model performance by concentrating more on minor data in the process of learning\cite{zhang2020novel, li2021autobalance}. Also, data augmentation methods have been studied for solving the imbalanced data problem\cite{dos2019data, jiang2020data, abaskohi-etal-2022-utnlp}.

\subsection{Feature Extraction}
\label{subsec:featureextraction}

Specific features of social media texts like emojis and hashtags usually carry valuable information\cite{alfano2021affiliative}. In English, \cite{sari2014user} and \cite{lemmens2020sarcasm} used these features to improve the result of their model by giving them separately to the model. This method has been used in Persian for rumor detection, and verification \cite{mahmoodabad2018persian, jahanbakhsh2021semi}. In this paper, in addition to using emojis and hashtags, we used Part of Speech(POS) and misspelled words. We believe misspelled words are usually used to grab more attention, which is a reason for having a piece of helpful information. Furthermore, as the language models used in our experiments do not follow the formal sentence structure used in the language models' pre-training phase, POS tagging helps the model better understand the sentence structure.

\section{Methods and Data}
\label{sec:method}

This section explains the properties of the dataset utilized in this work. In addition, we will contrast this dataset with other Persian datasets currently accessible for emotion recognition. Finally, we will outline our process to develop a superior generic model for emotion recognition.

\subsection{Datasets}
\label{subsec:dataset}

Emotion detection, as one of the complex tasks in natural language processing, not only has challenges in detecting tweets but also the data, as the vital part of the deep learning model, has its challenges.

EmoPars\cite{sabri-etal-2021-emopars-collection}, is a dataset of 30,000 Persian Tweets labeled with Ekman's six basic emotions (Anger, Fear, Happiness, Sadness, Hatred, and Wonder)\cite{ekman1992there}. The first publicly available Persian emotion dataset provides enough information for further data analysis. Although other available datasets like ArmanEmo\cite{mirzaee2022armanemo} believe each sentence can only contain one emotion, it seems emotion detection is a multi-label task\cite{bhowmick2009multi, deng2020multi, fei2021machine}. In contrast, the EmoPars dataset considers this task as a multi-label task. In this paper, we use multi-label training in one of our experiments on EmoPars; however, using multiple binary classifiers reaches better results since in this approach handling data imbalance is easier and the model has a simple task as well since it faces just two classes.

EmoPars uses information obtained from Twitter's official developer API\cite{sabri-etal-2021-emopars-collection}. They gather tweets using multiple keywords, including articles and prepositions, from various topics to avoid data being skewed by any topic or debate. Include at least a few tweets from each period in your tweets. As in the published dataset, the vote counts have been provided in the range of [0, 5] for classification; various policies can be used to create a classification (whether binary, multi-label, or multi-class). Here we try to find out the best policy for classification.


ArmanEmo\cite{mirzaee2022armanemo} is another dataset whose data is gathered from Twitter, Instagram, and customers’ comments on Digikala (an online shopping platform), and it consists of approximately 2.55 million raw samples(1.5 million from Twitter, 1 million from Instagram, and 50K from Digikala). Unlike EmoPars, they removed samples whose lengths felt outside of a specific range or included specific user IDs or links in the annotation step. Furthermore, they controlled the class balance and considered only using samples with one emotion and removed samples with multiple or no emotions.


\subsection{Data Selection Policy}
\label{subsec:datapolicy}

Data annotation is not a simple task, especially in emotion detection, where some emotions are debatable, and detecting whether they have a specific emotion is more challenging and may result in incorrectly labeled data. The voting method used in this paper's datasets, EmoPars and ArmanEmo, can cause some disagreement in the data. Although ArmanEmo tries to handle this problem by removing samples with no emotion or multiple emotions, EmoPars does not.


One possible solution might be using a threshold to annotate samples. Additionally, tweets close to this level can result in other issues. Using a specific threshold to decide whether a tweet contains an emotion can be helpful but finding this threshold may be challenging. In addition, tweets near this threshold may cause different problems.

Although we have tried different threshold values, and the results are discussed in Section \ref{sec:experiments}, we are introducing a new strategy for selecting valuable data for training our model. In this approach, the tweet is not considered when two or three individuals out of 5 annotators agree that a tweet contains a particular emotion.

We have also selected thirty tweets from each emotion's label and labeled them ourselves. The results in Table \ref{table:labeled-data} show us that our approach is acceptable. In general, there is no confidence in labels two and three. Nonetheless, about \textit{Wonder} emotion, our general idea is applicable; the crucial point here is the hardness of predicting \textit{Wonder} label. It is important to note that fewer than thirty tweets with the label "5" for each emotion. Because of this, we have included this label in all tweets.

\begin{table}[!ht]
    \centering
    \scriptsize
    \renewcommand{\arraystretch}{1}
    \begin{tabular}{lc|cccccc}
        \multicolumn{2}{c|}{\multirow{2}{*}{\textbf{Emotion}}} & \multicolumn{6}{c}{\textbf{Labels}} \\
        \multicolumn{2}{c|}{}                        & \textbf{0} & \textbf{1} & \textbf{2} & \textbf{3} & \textbf{4} & \textbf{5} \\ \hline
        \multirow{2}{*}{\textbf{Anger}}       & \texttt{\textbf{Predicted 0}}  & 23 & 21 & 17 & 15 & 9 & 3 \\
                                     & \texttt{\textbf{Predicted 1}}  & 7 & 9 & 13 & 15 & 21 & 21  \\ \hline
        \multirow{2}{*}{\textbf{Fear}}      & \texttt{\textbf{Predicted 0}}  & 24 & 23 & 18 & 16 & 10 & 1  \\
                                     & \texttt{\textbf{Predicted 1}}  & 6 & 7 & 12 & 14 & 20 & 1  \\ \hline
        \multirow{2}{*}{\textbf{Sadness}}   & \texttt{\textbf{Predicted 0}}  & 23 & 28 & 27 & 18 & 10 & 6  \\
                                     & \texttt{\textbf{Predicted 1}}  & 7 & 2 & 3 & 12 & 20 & 22  \\ \hline
        \multirow{2}{*}{\textbf{Happiness}}     & \texttt{\textbf{Predicted 0}}  & 28 & 25 & 18 & 16 & 12 & 1  \\
                                     & \texttt{\textbf{Predicted 1}}  & 2 & 5 & 12 & 14 & 18 & 10  \\ \hline
        \multirow{2}{*}{\textbf{Wonder}}      & \texttt{\textbf{Predicted 0}}  & 27 & 28 & 21 & 17 & 15 & 1  \\
                                     & \texttt{\textbf{Predicted 1}}  & 3 & 2 & 9 & 13 & 15 & 7  \\ \hline
        \multirow{2}{*}{\textbf{Hatred}}        & \texttt{\textbf{Predicted 0}}  & 22 & 20 & 15 & 16 & 10 & 1  \\
                                     & \texttt{\textbf{Predicted 1}}  & 8 & 10 & 15 & 14 & 20 & 12 
    \end{tabular}
    \caption{The result of labeling thirty tweets of each label of each emotion to check whether removing tweets with two or three agreed annotators is an acceptable approach.}
    \label{table:labeled-data}
\end{table}

\subsection{Data Augmentation}
\label{subsec:dataaugmentation}

\begin{figure*}[!ht]
    \centering
    \includegraphics[scale=0.24]{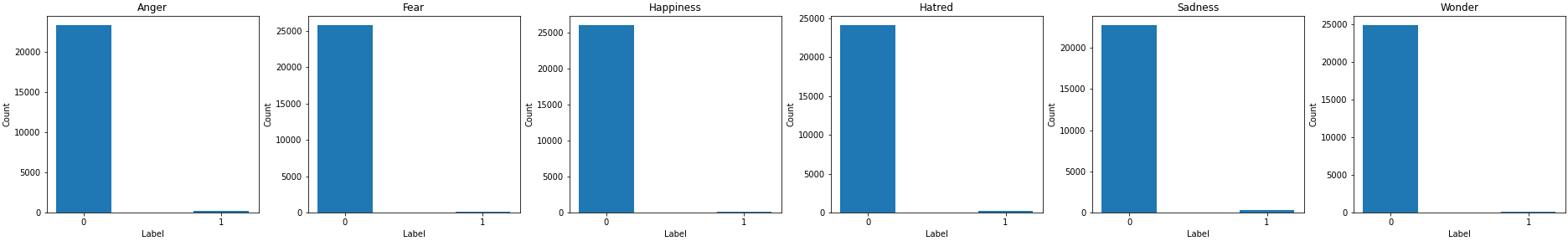}
    \caption{An overview of the number of data samples in each emotion. There are 1 to 100 different labels for each emotion. With this ratio, the model will predict 0 in most cases.}
    \label{fig:datacount}
\end{figure*}

One of the main challenges in solving emotion detection using the EmoPars dataset is the unbalanced data. 
Figure \ref{fig:datacount} shows a big difference between 0 and 1 labels in each emotion. This unbalanced data will result in biased results. There are several approaches to solve this issue.

\begin{figure}[!ht]
    \includegraphics[scale=0.55]{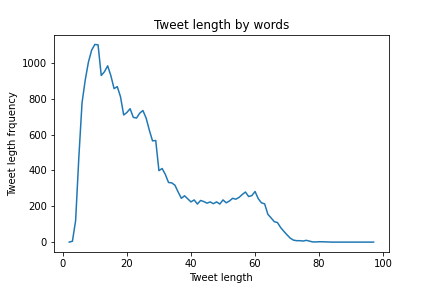}
    \caption{Length of each tweet by words. As most samples have less than 20 words, data augmentation would create another view of the initial tweet.}
    \label{fig:tweetlength}
\end{figure}

Although there is a massive difference in our case, it may not only not help us but also weaken the result, as was shown in\cite{aquino2017effect}. In addition, our analysis of the length of the tweets which can be seen in Figure \ref{fig:tweetlength}, shows us that creating 100 new views of the initial sample may create tweets in which the sentiment of the sentence is different or tweets with the same sentiment.

Mutation-based data augmentation, which uses synonym replacement, word elimination, word addition, and word swap, has been shown to improve the models' results in various tasks\cite{kafle2017data, parida2019abstract, shorten2021text, abaskohi-etal-2022-utnlp}. We used this data augmentation to create more samples of label 1 of each emotion to decrease the difference between classes.

In our data augmentation, we used NLPAug\cite{ma2019nlpaug} python module's ContextualWordEmbsAug and RandomWordAug methods. This module uses a language model for word insertion and synonym replacement. We used ParsBERT\cite{ParsBERT} as the language model in this section. Due to the different ratios between 0 and 1 classes in each emotion, the number of augmented samples per initial sample is different per emotion. Table \ref{table:augmentation-info} shows data augmentation hyper-parameters used for each emotion.

\begin{table*}[!ht]
    \normalsize
    \centering
    \renewcommand{\arraystretch}{1}
    \begin{tabular}{c|cccc|c}
        \multirow{2}{*}{\textbf{Emotion}} & \multicolumn{4}{c|}{\textbf{Hyper-parameters}}                                        & \multirow{2}{*}{\textbf{\# Augmented Samples}} \\
                                          & \textbf{Swap p} & \textbf{Replacement p} & \textbf{Insertion p} & \textbf{Deletion p} &                                                       \\ \hline
        \textbf{Anger}                    & 0.6             & 0.6                    & 0.3                  & 0.3                 & 10                                                    \\
        \textbf{Fear}                     & 0.6             & 0.6                    & 0.5                  & 0.4                 & 20                                                    \\
        \textbf{Happiness}                & 0.6             & 0.6                    & 0.5                  & 0.4                 & 20                                                    \\
        \textbf{Hatred}                   & 0.6             & 0.6                    & 0.3                  & 0.3                 & 10                                                    \\
        \textbf{Sadness}                  & 0.6             & 0.6                    & 0.3                  & 0.3                 & 10                                                    \\
        \textbf{Wonder}                   & 0.6             & 0.6                    & 0.4                  & 0.4                 & 15                                                   
    \end{tabular}
    \caption{Hyper-parameters for mutation-based data augmentation are used in the approach. The number of augmented samples per initial sample varies for each emotion as the ratio between the number of samples in each class is different. "p" here stands for probability.}
    \label{table:augmentation-info}
\end{table*}

\subsection{Model}
\label{subsec:model}

This section presents baseline models for EmoPars transfer learning-based emotion classification. Since labels are expensive, supervised learning is challenging to apply to NLP issues, including emotion detection. Transfer learning is helpful in this situation. In recent years, state-of-the-art performance in various NLP tasks has resulted from transfer learning from pre-trained deep neural Language Models (LMs) towards downstream language issues\cite{ying2019improving, durrani2021transfer}.

We use the ParsBERT\cite{ParsBERT} pre-trained language model for Persian in most of our experiments to find the best solution. A monolingual language model called ParsBERT is built on the BERT architecture\cite{devlin-etal-2019-bert}. The ParsBERT model surpasses the multilingual BERT\cite{pires-etal-2019-multilingual} and earlier models in some Persian NLP downstream tasks, such as text categorization and sentiment analysis, according to research by Farahani et al.\cite{ParsBERT}. ParsBERT has been trained on a broader and more varied set of pre-trained Persian datasets than the multilingual BERT, making it lighter in weight. Although, due to the unbalanced data, we used other strategies to increase the model's performance. We will introduce them in the following sections.

In addition, to compare the quality of the EmoPars and ArmanEmo (a human-labeled emotion dataset of more than 7000 Persian sentences labeled for seven categories)\cite{mirzaee2022armanemo}, we used XLM-RoBERTa\cite{conneau-etal-2020-unsupervised} and XLM-EMO\cite{bianchi-etal-2022-xlm}.

A multilingual language model called XLM-RoBERTa has already been trained using Masked Language Modeling (MLM). XLM-RoBERTa performs better than multilingual BERT since it has more training data. XLM-RoBERTa is trained on 100 languages was.

A multilingual emotion prediction model called XLM-EMO was developed using social media data. The model performs admirably in a zero-shot scenario, indicating that it is helpful for low-resource languages.

\subsubsection{Data Pre-processing}
\label{subsubsec:datapreprocessing}

The data pre-processing method used in our approach was like what was used in ArmanEmo paper\cite{mirzaee2022armanemo}. We used the Parsivar toolkit\cite{mohtaj-etal-2018-parsivar} in this manner. Along with various character refining procedures, Parsivar does several rule-based space correction operations (including word, punctuation, and affix spacing) (such as removing stretching letters). However, its normalization guidelines are not exhaustive. In addition to normalization, we did the following pre-processings:

\begin{itemize}
    \item {If there is an English word in the sentence; first we check if the word exists in a single token to single token dictionary provided by Meta\footnote{It is available at \url{https://dl.fbaipublicfiles.com/arrival/dictionaries.tar.gz}}. If the word does not exist in the dictionary, we use transliteration provided by Polyglot module\footnote{\url{https://polyglot.readthedocs.io/en/latest/polyglot.html}}. Information from other languages is not considered in our approach.}
    \item {Removing letters from Persian words that are frequently purposely misspelled in order to emphasize certain words more strongly in casual writing.}
    \item {Removing from the text any Arabic diacritics that the Parsivar normalizer did not remove.}
    \item {After completing the aforementioned stages, we removed any last non-Persian characters.}
    \item {Removing the "\#" from the text's hashtags while retaining the information they contain.}
\end{itemize}

\subsubsection{Loss Function}
\label{subsubsec:lossfunction}

The loss function greatly impacts a model's performance. The influence of the different loss functions has been investigated in \cite{abdel-salam-2022-reamtchka} for the sarcasm detection task. In their results, they found that F1-Cross-Entropy and Recall-Cross-Entropy may help that to reach a good F1-Sarcastic. Equation \ref{equation:f1loss} and \ref{equation:recallloss} shows how this loss functions work respectively. In these equations, $F1_{c}$ and $Recall_{c}$ are the F1-Score and Recall corresponding to a specific class c, respectively.

\begin{equation}
    \centering
    \label{equation:f1loss}
    \sum_{c=1}^{C} -(1-F1_{c}) \times N_{c} \times \log(P_{c})
\end{equation}

\begin{equation}
    \centering
    \label{equation:recallloss}
    \sum_{c=1}^{C} -(1-Recall_{c}) \times N_{c} \times \log(P_{c})
\end{equation}

Although in the mentioned task, Recall-Cross-Entropy reaches better results, in our task, F1-Cross-Entropy showed better results.

\subsubsection{Feature Extraction}
\label{subsubsec:featureextraction}

\begin{figure*}[!ht]
    \includegraphics[scale=0.42]{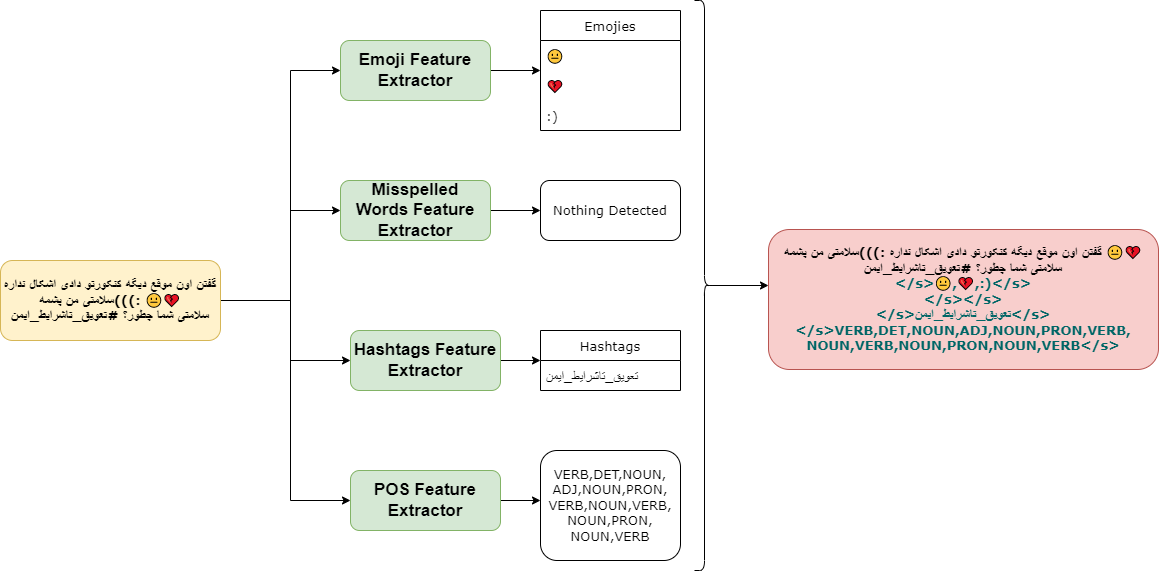}
    \caption{We use different feature extractors, and then we add the extracted features to the pre-processed tweet using the "</s></s>" token. The input sentence translation in English is: "There's nothing wrong with saying that you took the exam at that time :))). My health does not matter <broken-heart-emoji> <neutral-face-emoji>. How about yours?\#postponement\_of\_safe\_conditions".}
    \label{fig:featureextraction}
\end{figure*}

Although considering the whole context gives the sample's sentiment and emotion, some specific parts of the samples, like hashtags or emojis, include more information. Considering these features, as well as the embedding of the sample, showed improvement in previous studies\cite{hernandez2015applying, yaghoobian2021sarcasm, nandwani2021review}. After extracting some lexical-based features, using the splitting token "\textless/s\textgreater\textless/s\textgreater", all text features are simply added to the pre-processed tweets. Figure \ref{fig:featureextraction} shows how this process works.

The feature extraction methods we used for this task are:

\begin{itemize}
    \item {Emoji: After examining a sizable amount of Twitter data, emoji are a notable multi-model feature that depicts human emotions.}
    \item {Parts of Speech(POS): Additionally, changing the sentence structure sometimes is used to show the writer's emotion. As a result, POS tags are utilized as a crucial component.}
    \item {Misspelled Words: Occasionally, incorrectly spelled and capitalized words that are not at the start of a phrase might increase the emotion.}
    \item {Hashtags: Hashtags show the main point of the sentence more clearly. Using them as a particular feature would be helpful.}
\end{itemize}

\subsubsection{Undersampling and Class Weight}
\label{subsubsec:undersampling}

In Section \ref{subsec:dataaugmentation}, we used data augmentation to decrease the big difference between the number of samples between the two classes. In addition, we mentioned that due to the vast difference, creating nearly 100 views of a sentence using augmentation methods not only does not help us but may also weaken the performance of the model. Other methods can help us to do, in addition to data augmentation, to make the number of overcome the unbalanced dataset problem.

Instead of increasing the number of samples for the minor class, we can decrease the number of samples in the significant class, which is called undersampling. Undersampling is a technique to balance uneven datasets by keeping all of the data in the minority class and decreasing the size of the majority class. This method has been investigated in previous papers\cite{zhu2007active, kapadi2022natural}. To do undersampling, we selected the tweets randomly with a preference for shorter tweets(using $\frac{1}{Tweet Length}$ as the weight of our selection).

In addition to undersampling, class weights can be used in the training approach. Most machine learning algorithms are not very useful with partial class data. However, we can modify the current training algorithm to consider the skewed distribution of the classes. This can be achieved by giving weight to both the majority and minority classes. The difference in weights will influence the classification of the classes during the training phase. The purpose is to penalize the misclassification made by the minority class by setting a higher class weight and simultaneously reducing the weight for the majority class.

We decreased our dataset imbalance using data augmentation, undersampling, and class weights. Table \ref{table:finaldataset} shows information about the final dataset to the model. In addition to these methods, feature extraction introduced in Section \ref{subsubsec:featureextraction} and the F1-Cross-Entropy loss function mentioned in Section \ref{subsubsec:lossfunction} were also helpful.

\begin{table}[!ht]
    \small
    \centering
    \renewcommand{\arraystretch}{1}
    \begin{tabular}{l|cc|c}
        Emotion   & \# of 0 Samples & \# of 1 Samples & Weights \\ \hline
        Anger     & 2590                & 13986               & 5-1           \\
        Fear      & 1520                & 12768               & 8-1           \\
        Happiness & 1440                & 14562               & 9-1           \\
        Hatred    & 1750                & 12263               & 7-1           \\
        Sadness   & 2990                & 13643               & 4-1           \\
        Wonder    & 1935                & 14512               & 7-1          
    \end{tabular}
    \caption{The information of the final dataset is provided to the model. Class weights are reported in "weight of the 0 class-weight of the 1 class".}
    \label{table:finaldataset}
\end{table}

\section{Experiments}
\label{sec:experiments}

In this section, we report our results in different experiments. We compare two famous Persian emotion datasets with our approach. In addition, on EmoPars, we evaluate the effectiveness of our method through different experiments, including multi-label training and different threshold in selecting tweets.

In our experiments, 10\% of the datasets have been used for evaluation and the rest for training. In addition, batch size 16, 5 epochs, the learning rate of 1e-4, and a max length of 256 were utilized in these experiments.

\subsection{Training With Using Class Weights Only}
\label{subsec:multi-label}

Because no baseline has been provided in the EmoPars paper and the analysis in the ArmanEmo paper on this dataset is not fair, we first trained ParsBERT without using any of the other methods. The only method that has been used to handle imbalanced datasets was using class weights. The class weight was set based on the number of samples, where the ratio of the number of significant class samples by the number of minor class samples was used as the weight of the minor class, and the weight of the major class was one.

\begin{table}[]
    \centering
    \small
    \begin{tabular}{l|cccc}
    Emotion   & Accuracy & Precision & Recall & F1-score \\ \hline
    Anger     & 0.38     & 0.15      & 0.39   & 0.21     \\
    Fear      & 0.49     & 0.24      & 0.49   & 0.32     \\
    Happiness & 0.51     & 0.26      & 0.51   & 0.34     \\
    Hatred    & 0.42     & 0.17      & 0.42   & 0.25     \\
    Sadness   & 0.38     & 0.15      & 0.38   & 0.21     \\
    Wonder    & 0.45     & 0.2       & 0.45   & 0.28                     
    \end{tabular}
    \caption{Baseline results for evaluating the effectiveness of our approach. Class weights are used to reach good results for this result in addition to data preprocessing.}
    \label{table:baselineres}
\end{table}

Table \ref{table:baselineres} shows the baseline results for EmoPars dataset. The interesting fact about these results might be that although happiness has the best results, sadness does not show this fact. This is because people use more specific words in happiness, while in sadness, the words can conflict with those in hatred. Another reason may be the larger number of samples in class zero in sadness than happiness, resulting in using a larger class weight for class one.

\subsection{Multi-label Training}
\label{subsec:multi-label}

As mentioned in Section \ref{subsec:dataset}, emotion detection is multi-label task which EmoPars considers this fact. One of the first experiments we did was using the ParsBERT model for multi-label training of the model. We used Hamming Loss to evaluate the model. Equation \ref{equation:hl} shows how hamming loss is calculated. In this formula, $N$ is the number of samples, $L$ is the number of labels, and $\oplus$ denotes exclusive-or. 

\begin{equation}
    \centering
    \label{equation:hl}
    HL = \frac{1}{N L} \sum_{l=1}^L\sum_{i=1}^N Y_{i,l} \oplus X_{i,l}
\end{equation}

\begin{table}[]
    \centering
    \begin{tabular}{c|cc}
    \multicolumn{1}{c|}{\multirow{2}{*}{Threshold}} & \multicolumn{2}{c}{Test Result}                   \\
    \multicolumn{1}{c|}{}                           & \multicolumn{1}{c}{Hamming Loss} & Hamming Score \\ \hline
    1                                               & \multicolumn{1}{c|}{0.4653}       & 0.4022        \\
    2                                               & \multicolumn{1}{c|}{0.1848}       & 0.3265        \\
    3                                               & \multicolumn{1}{c|}{0.0383}       & 0.803         \\
    4                                               & \multicolumn{1}{c|}{0.0054}       & 0.9685        \\
    5                                               & \multicolumn{1}{c|}{0.0005}       & 0.9971       
    \end{tabular}
    \caption{The result of multi-label training with ParsBERT. The results are reported based on different thresholds—the tweets with votes more equal to or more than the threshold considered having the emotion.}
    \label{table:multilabelresult}
\end{table}

\begin{figure*}[!ht]
    \centering
    \includegraphics[scale=0.43]{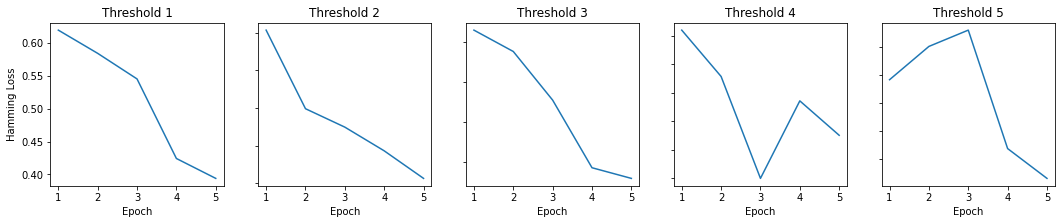}
    \caption{An overview of hamming loss of ParsBERT that was trained on the EmoPars per epoch.}
    \label{fig:hammingtreshbatch}
\end{figure*}

Table \ref{table:multilabelresult} shows the result of multi-label training with different thresholds on ParsBERT. The hamming score shows the accuracy of the model. Although we have a high score on the last two thresholds, we should consider this result unreliable since we face imbalanced data, which Figure \ref{fig:datacount} depicts. When the considered threshold is bigger, the number of samples in one class(class zero) is higher. As a result, we have a more hamming score. Even though multi-label is more efficient, we have to use multiple binary classifiers to detect the emotions in one sample. Moreover, Figure \ref{fig:hammingtreshbatch} illustrates loss per each batch for every five epochs which is less smooth for thresholds 4 and 5. 

\subsection{Removing Indeterminate Data}
\label{subsec:indeterminateexp}

One of our ideas to evaluate the EmoPars dataset properly was removing indeterminate data explained in Section \ref{subsec:datapolicy}. Before applying our method for handling imbalanced data, we need to check the effect of the provided policy.

\begin{table}[]
    \small
    \centering
    \begin{tabular}{l|ccccc}
    Emotion   & Precision & Recall & F1-score & Support \\ \hline
    Anger     & 1         & 0.25   & 0.33     & 2357    \\
    Fear      & 0.43      & 0.31   & 0.36     & 2785    \\
    Happiness & 0.28      & 0.44   & 0.34     & 2615    \\
    Hatred    & 0.57      & 0.2    & 0.29     & 2425    \\
    Sadness   & 0.76      & 0.22   & 0.35     & 2305    \\
    Wonder    & 0.57      & 0.17   & 0.27     & 2501   
    \end{tabular}  
    \caption{\label{table:indeterminateremoval}The result of ParsBERT on the EmoPars after deleting samples with labels two and three and using class weights. Support shows the number of samples used for evaluation. Accuracy in all emotions was 0.99 except for Happiness which was 0.98.}
\end{table}

Table \ref{table:indeterminateremoval} shows the effect of removing samples with labels two and three. As the results show, F1-score has been increased to 0.1, which is a noticeable improvement in this imbalanced dataset.

\subsection{Impact Of Our Approach}

Since we were facing with imbalanced data, we used some approaches to solve this issue which is introduced in Sections \ref{subsec:dataaugmentation}, \ref{subsubsec:datapreprocessing}, \ref{subsubsec:lossfunction}, \ref{subsubsec:featureextraction}, and \ref{subsubsec:undersampling}. The combination of all of these methods resulted in significant improvement in the result of our method.

\begin{table}[]
    \centering
    \small
    \begin{tabular}{l|ccccc}
    Emotion   & Accuracy & Precision & Recall & F1-score \\ \hline
    Anger     & 0.81     & 0.7         & 0.78   & 0.74   \\
    Fear      & 0.79     & 0.65      & 0.81   & 0.72     \\
    Happiness & 0.81     & 0.68      & 0.84   & 0.75     \\
    Hatred    & 0.83     & 0.75      & 0.74    & 0.75     \\
    Sadness   & 0.87     & 0.81      & 0.81   & 0.81     \\
    Wonder    & 0.86     & 0.8      & 0.78   & 0.79        
    \end{tabular}
    \caption{\label{table:oursresults}The results of ParsBERT on the EmoPars using F1-Cross-Entropy, Data Augmentation, Undersampling, Class Weights, and Feature Extraction.} 
\end{table}

\begin{table*}[]
    \centering
    \begin{tabular}{l|cccccc|c}
    \multirow{2}{*}{Model} & \multicolumn{6}{c|}{Test F1-score}                   & \multirow{2}{*}{Macro-F1} \\
                           & Hatred & Happiness & Sadness & Wonder & Anger & Fear &                           \\ \hline
    XLM-EMO                & 0.78   & 0.77      & 0.82    & 0.82   & 0.8   & 0.75 & 0.79                      \\
    XLM-RoBERTa            & 0.79   & 0.79      & 0.84    & 0.85   & 0.81  & 0.79 & 0.81                      \\
    ParsBERT               & 0.74   & 0.72      & 0.81    & 0.79   & 0.74  & 0.72 & 0.76                     
    \end{tabular}
    \caption{\label{table:comparisonmodels}Macro-averaged F1-score and each emotion F1-score for different models on EmoPars.} 
\end{table*}

Table \ref{table:oursresults} shows our approach has approximately double the F1-score in each emotion. We believe this result is due to using a small amount of each method for handling the imbalanced data. Using just data augmentation can make redundant samples, and under-sampling can decrease the number of samples. Nevertheless, using both of them and adding more attention by feature extraction can help us more. Moreover, the F1-Cross-Entropy tries to increase the F1-score instead of Accuracy, which will help us to reach higher precision and recall.

\begin{figure*}[!ht]
    \centering
    \includegraphics[scale=0.33]{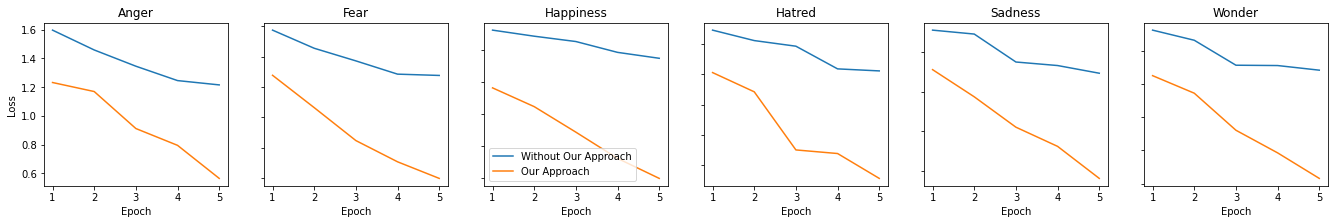}
    \caption{An overview of the loss value while employing our strategy and a comparison to training the model without it.}
    \label{fig:ourvsnon}
\end{figure*}

\begin{table}[]
    \centering
    \begin{tabular}{l|cc}
    \multirow{2}{*}{Train Set} & \multicolumn{2}{c}{Test Set} \\
                               & EmoPars      & ArmanEmo      \\ \cline{2-3} \hline 
    Arman Emo                  & 0.76         & 0.81          \\
    EmoPars                    & 0.78         & 0.74         
    \end{tabular}
    \caption{\label{table:comparison}Macro-averaged F1-score for comparison between ArmanEmo and EmoPars datasets.} 
\end{table}

Figure \ref{fig:ourvsnon} shows the influence of our approach on training. Feature extraction has a significant impact on the initial loss of the model. Then, the loss function helps the model. In addition, since the model sees more samples in the positive class, it can reach better results.

\subsection{Comparing Datasets}
\label{subsec:comparing}

One of the experiments in \cite{mirzaee2022armanemo} compared the provided dataset, ArmanEmo, and the EmoPars dataset. However, their data selection policy was not fair and accurate. They just used 919 samples from EmoPars, a small amount compared to 4700 samples in ArmanEmo.

Here, we trained the model on EmoPars with more samples and used the same model and imbalanced data handling approaches. Table \ref{table:comparison} shows the results of our comparison. Unlike the results provided in ArmanEmo, training on one dataset results in a better F1-score on the same dataset. However, results show that EmoPars is noisier and harder to learn, which can be because of including multiple emotions in one sample.

\subsection{Comparison Among Different Models}
\label{subsec:comparisonmodels}

Another experiment that we have done was evaluating different models on EmoPars. All the experiments till this section were done using the ParsBERT model. Table \ref{table:comparisonmodels} shows the results of the three models we have used: XLM-EMO\cite{bianchi-etal-2022-xlm}, XLM-RoBERTa\cite{bianchi-etal-2022-xlm}, and ParsBERT\cite{ParsBERT}. XLM-RoBERTa and XLM-EMO have higher values due to having a higher number of parameters.

\section{Conclusion}
\label{sec:conclusion}

This paper uses an approach for handling imbalanced data on EmoPars, an emotion recognition dataset. Since EmoPars produces imbalanced multi-label data labeled with voting technique, selecting definite data is essential. Our experiment on EmoPars shows that removing uncertain samples increased the F1-score of the method up to 51\%. However, this could not help with imbalanced data, which results in a biased model. As a result, we used data augmentation (including swapping words, synonym substitution, word deletion, and word addition), undersampling, and F1-Cross-Entropy loss function to solve this issue. In addition, we used a feature extraction technique that adds emojis, hashtags, POS tags, and misspelled words at the end of the sentence to improve the performance. Our model reaches a Macro-F1-score of 0.76 on the EmoPars. Furthermore, our introduced method reaches a Macro-F1-score of 0.81 on ArmanEmo\cite{mirzaee2022armanemo}.

\bibliography{anthology,custom}
\bibliographystyle{acl_natbib}

\end{document}